\documentclass[11pt,conference]{article}
\usepackage{cite}
\usepackage{amsmath,amssymb,amsfonts}
\usepackage{algorithmic}
\usepackage{graphicx}
\usepackage{textcomp}
\usepackage{xcolor}
\usepackage{wrapfig}
\usepackage{INTERSPEECH2019}
\usepackage{paralist}

\def\BibTeX{{\rm B\kern-.05em{\sc i\kern-.025em b}\kern-.08em
    T\kern-.1667em\lower.7ex\hbox{E}\kern-.125emX}}

\title{Automatic Spoken Language Identification using a Time-Delay Neural Network}
\name{Benjamin Kepecs$^1$ and Homayoon Beigi$^2$}
\address{
  $^1$Columbia University\\
  $^2$Recognition Technologies, Inc. and Columbia University}
\email{$^1$bek2127@columbia.edu, $^2$beigi@recotechnologies.com}

\begin{document}

\maketitle

\begin{abstract}
Closed-set spoken language identification is the task of recognizing
the language being spoken in a recorded audio clip from a set of known
languages. In this study, a language identification system was built
and trained to distinguish between Arabic, Spanish, French, and
Turkish based on nothing more than recorded speech. A preexisting
multilingual dataset was used to train a series of acoustic models
based on the Tedlium TDNN model to perform automatic speech
recognition. The system was provided with a custom multilingual
language model and a specialized pronunciation lexicon with language
names prepended to phones. The trained model was used to generate
phone alignments to test data from all four languages, and languages
were predicted based on a voting scheme choosing the most common
language prepend in an utterance. Accuracy was measured by comparing
predicted languages to known languages, and was determined to be very
high in identifying Spanish and Arabic, and somewhat lower in
identifying Turkish and French.
\end{abstract}

\section{Introduction}
Spoken language identification involves recognizing a language based
on a snippet of recorded speech. There are multiple potential uses for
this technology. Real-time automatic translation and transcription
systems often require prior information about the input language in
order to operate, and would not be suitable in an environment where
multiple languages are spoken at unpredictable times, such as at a
global conference. Automatic spoken language identification can be
used at the beginning of a multilingual speech recognition pipeline to
obviate the need for human input and make the software work more
seamlessly. In particular, spoken language identification could play a
role in a code-switching detection pipeline to keep track of when
someone has begun to speak in a new language, and what language they
are speaking in. Another potential application of this technology is
in customer service; currently, to serve diverse global clients a
customer service hotline may automatically list several languages and
direct the caller to press a button corresponding to their
language. Automatic spoken language identification could be used to
detect a caller's language and automatically route their call to a
native speaker or list additional options in their language.\medskip

\section{Related Works}
There have been many research studies exploring ways to implement
spoken language identification using both neural networks and more
traditional acoustic models. There are, in addition, several basic
language identification utilities built into commercial products such
as Google Cloud, Microsoft Azure, and
AWS~\cite{google,eric-urban,engdahl_2008}. These technologies, however,
are still in their early stages of application; automatic language
identification was only added to AWS Transcribe in late 2020.\medskip

One of the first semi-successful approaches to language ID was to use
a panel of Hidden Markov Models (HMMs), each trained on a single
language~\cite{zissman_1993}. Speech of some unknown language would
then be decoded with each of the HMMs in turn, and the language of the
model which decoded the speech most accurately was said to be the
unknown language. This method was further refined by using separate
stochastic models for each phoneme of the target languages and
decoding speech with a distinct series of phoneme models for each
language~\cite{muthusamy_barnard_cole_1994}. In some cases, these
models took into account prosodic elements by manually adjusting
features to incorporate prosodic characteristics of the
languages~\cite{hazen_zue_1997,muthusamy_cole_gopalakrishnan_1991}.\medskip

More recent works on automatic language ID use deep-learning based
methods to train neural networks as acoustic models. One approach is
to treat the problem of language identification as a computer vision
classification problem, and thus to train a CNN on spectrograms
labeled with their corresponding
languages~\cite{singh_sharma_kumar_kaur_baz_masud_2021}. This approach,
though it meets with some success in limited scenarios, must generate
unnecessary intermediate images and disregards decades of progress on
acoustic feature extraction and acoustic model generation.\medskip

Other state-of-the-art deep-learning approaches are more in line with
traditional ASR and use convolutional neural networks (CNNs) or
regular deep neural networks (DNNs) as acoustic models, accepting
labeled i-vectors as features~\cite{ieee_dl_comparison}. These methods
achieve 3.48\% and 3.55\% equal error rates (EER), respectively, on language
identification among 50 different languages. Surprisingly, integrating
two different classifiers such as a CNN and a Support Vector Machine
(SVM) by running them in parallel and then adding the resulting scores
yields an even lower EER of 2.79\%. Though Google,
Inc. does not disclose the technology that drives it automatic
language identification on Google Cloud, it is likely that a DNN-based
method with i-vector features is used in their application as
well~\cite{google_2014}.\medskip

\section{Dataset}
The training data used in this study was the MediaSpeech
dataset~\cite{mediaspeech2021}, which contains approximately 10 hours
of speech with matched transcriptions for Arabic, Spanish, French, and
Turkish. This multilingual dataset was chosen so that the data for all
four languages would have some degree of inter-language consistency in
audio quality and file format. The data was primarily derived from
European news channels, so it can be inferred that the French accents
represented in the dataset were France-accented French as opposed to
African-accented French. To test the system's performance, five
additional datasets were used: a dataset of female speakers in
Colombian Spanish~\cite{guevara-rukoz-etal-2020-crowdsourcing},
recited speech in Tunisian Modern Standard Arabic~\cite{tunisian_msa},
a Turkish Daily Use Sentence dataset~\cite{datasets_2022}, the
att-HACK French Expressive Speech Database with Social
Attitudes~\cite{lemoine:hal-02508362}, and finally a dataset of
African accented French~\cite{african}. The first four datasets were
used to test language ID performance on speech similarly accented to
the speech found in the MediaSpeech dataset, and the final African
accented French dataset was used to determine the robustness of the
prediction on unfamiliar accents.\medskip

\section{Methods}
\begin{figure}
\begin{minipage}[b]{1.0\linewidth}
\centering
\includegraphics[width=3in]{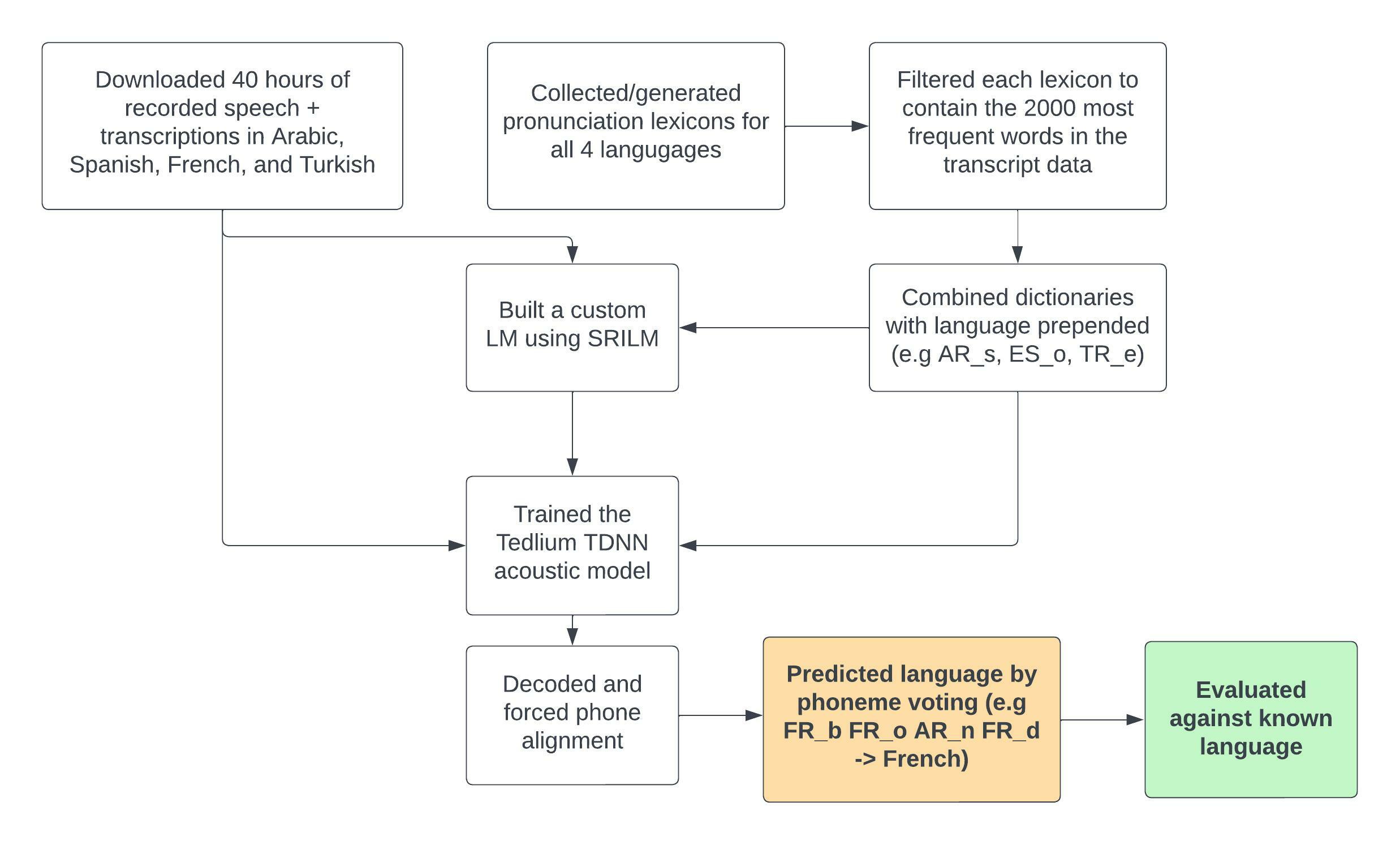}
\caption{High-level methodology for spoken language identification used in this study.}
\label{fig:speech_rec}
\end{minipage}
\end{figure}

Different languages have distinct phonological compositions,
vocabularies, and prosodies, and as such contain many plausible
features that could be used to distinguish one from another. The
approach presented in this study adapts the Kaldi Tedlium s5\_r3 egs
recipe for speech
recognition~\cite{Hernandez_2018,peddinti_povey_khudanpur_2015} to
perform language identification. Traditional MFCC feature
extraction~\cite{r:beigi-sr-book-2011} and acoustic modeling was
followed with refined modeling via a time-delay neural network. The
resulting trained network was used to decode testing data and
determine its language. A flowchart displaying the high-level steps in
this process is shown in Figure~\ref{fig:speech_rec}.\medskip

\subsection{Data and Lexicon Preparation}
The most recent version of Kaldi~\cite{povey11thekaldi} was cloned
from Github and compiled on a GCP VM instance. The MediaSpeech dataset
was then downloaded from OpenSLR, and all audio files were converted
to .wav from .flac formats. The data directory for Kaldi feature
extraction was constructed according to Kaldi specification.\medskip

\begin{figure}
  \begin{minipage}[b]{1.0\linewidth}
  \centering
  \includegraphics[width=3in]{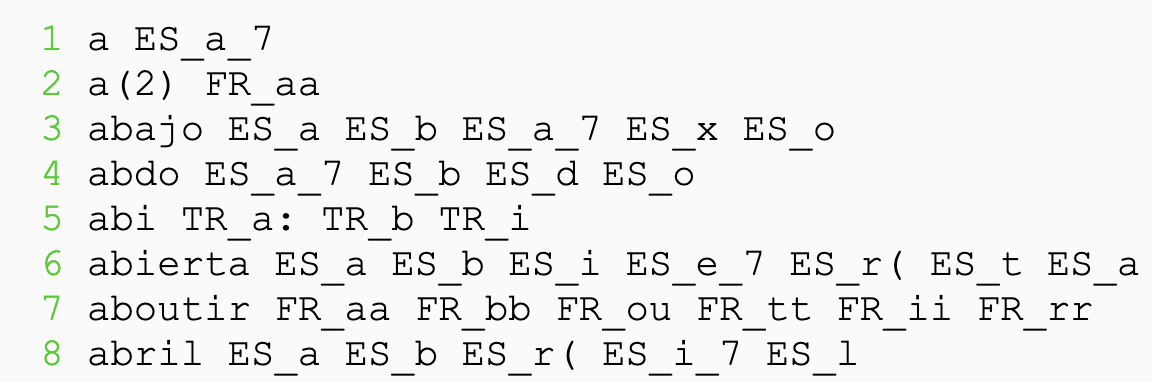}
  \caption{A small portion of the combined multilingual dictionary with language prepends and renumbering of different words with identical spellings.}
  \label{fig:dictionary}
  \end{minipage}
\end{figure}

A pronunciation lexicon for each language in the dataset was either
downloaded or manually generated: the French and Spanish pronunciation
lexicons were sourced from CMUDict~\cite{sphinx_2022}, the Turkish
lexicon was directly downloaded from a private Github
repository~\cite{turkish}, and the Arabic lexicon was generated by
feeding the combined corpus of Arabic utterances in the MediaSpeech
dataset into a Python tool for generating Arabic pronunciation
lexicons from a specified corpus~\cite{ardict}. The lexicon for each
language was filtered to contain only (at most) the 2000 most
frequently occurring words in the MediaSpeech transcriptions for that
language. To adjust the lexicon for the specific task of language
identification, the phones in each dictionary were prepended with a
symbol indicating the language of which they were a part. For example,
the 'e' phone in Spanish was prepended with 'ES', the 'aa' phone in
French was prepended with 'FR', etc. The dictionaries were combined
and words with duplicate spellings but different pronunciations among
and within languages were appropriately renumbered, as shown in
Figure~\ref{fig:dictionary}.\medskip

Next, the transcriptions provided for each language in the MediaSpeech
dataset were combined in random order into a single multilingual
corpus. This corpus was fed into SRILM~\cite{srilm} with order 4 to
generate a 4-gram multilingual language model in ARPA format. This was
then converted into an OpenFST format.\medskip

\subsection{Feature Extraction and Training}
As in the Tedlium egs recipe, feature extraction was performed on the
audio input to yield MFCC features for all recordings. 10,000 segments
in the dataset were subsetted, and these were used to perform flat
start monophone training. The monophones were aligned to
transcriptions, and the results were used to train delta-based
triphones. These too were aligned, and the results were used to
bootstrap and train LDA-MLLT triphones. Finally, the results were
aligned again and used to train SAT triphones.\medskip

The TDNN at the core of ASR acoustic modelling in the Tedlium recipe
is trained with iVectors. To generate iVectors for the training data,
the training data was speed-perturbed and volume perturbed and and
FMLLR-aligned to generate low-resolution and high-resolution perturbed
MFCC features. These components were used to train an iVector
extractor for the perturbed training and native testing
data. Alignment lattices and a decision tree were generated from the
low-resolution MFCCs, and subsequently the MFCCs, decision tree, and
extracted iVectors were used to train a 16-layer TDNN. The training
was performed on a NVIDIA Tesla T4 GPU. A new graph was generated and
the training data was decoded based on the high-resolution features
computed previously.\medskip

\subsection{Test Data Preparation}
Rather than withhold a portion of the MediaSpeech training data as
testing data, five additional datasets were used to construct a
testing partition. Aside from the practical considerations governing
this decision (namely, an oversight on behalf of the author), using
other datasets to test the model provided a more robust evaluation of
its performance in the real world because it introduced more
variability into the data quality, recording apparatus, and speaker
set. In addition to testing the performance of the model on
identifying languages similarly accented to the training data, the
robustness of the model in predicting the language of speech with an
out-of-training accented was also evaluated. The Colombian Spanish,
Tunisian Arabic, Expressive French, Daily Use Turkish, and African
Accented French datasets were downloaded and cleaned to remove
punctuation and capitalization. The data directories for each language
were assembled according to Kaldi specifications as before, and then
were subsequently combined into one multilingual testing data
directory. High-resolution MFCC features were computed for the
assembled testing data, and iVectors were extracted using the iVector
extractor trained previously.\medskip

\subsection{Decoding and Scoring}
\begin{figure}
  \begin{minipage}[b]{1.0\linewidth}
  \centering
  \includegraphics[width=3in]{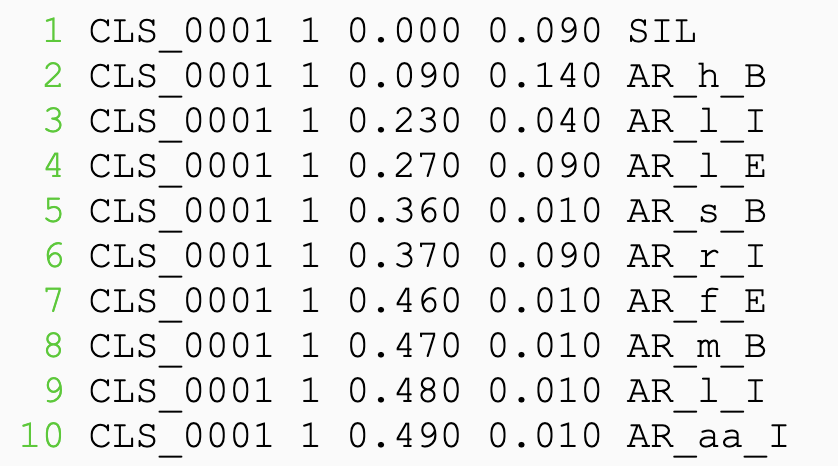}
  \caption{A sample of the phone symbol predictions produce by the decoding. The first column indicates the utterance name, the third and fourth columns indicate the time range, and the fifth column indicates the predicted phone symbol. Confidence values are excluded.}
  \label{fig:ctm}
  \end{minipage}
\end{figure}

The combined testing data directory was decoded using the graph
produced by the TDNN training stage. The decoding was used to generate
phone aligned lattices, and finally a series of phone number
predictions for each utterance organized by timestamp. Each phone
number was converted to a phone symbol by referencing the canonical
phone list used in training and decoding. A sample of the intermediate
.ctm file containing these phone symbol predictions is shown in
Figure~\ref{fig:ctm}.\medskip

Finally, the aligned phone predictions were used to assign a
hypothesized language ID to each utterance. The predicted phones for
each utterance were parsed to extract their prepended language
symbols, and a tally was maintained to keep track of the number of
phones of a particular language were present in an utterance; e.g, the
word "basura" with phone predictions ES\_b, ES\_a, FR\_s, FR\_u,
ES\_r, AR\_a would have a tally of ES: 3, FR: 2, AR: 1. Then, by a
simple voting scheme, the utterance was predicted to be the language
with the greatest tally. This prediction was compared to the known
language of the utterance in each case, and the prediction accuracy
rate was computed as the primary metric for assessing
performance.\medskip

\section{Results and Discussion}
\begin{figure}
  \begin{minipage}[b]{1.0\linewidth}
  \centering
  \includegraphics[width=3in]{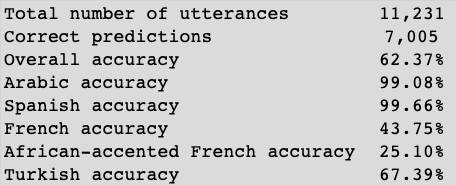}
  \caption{Results of decoding and scoring on the full testing datasets for Arabic, Spanish, French, African-accented French, and Turkish. The overall accuracy was computed by simply dividing the number of correct predictions by the total number of utterances. The accuracy for each specific language was computed by dividing the number of correct predictions for that language by the total number of utterances in that language.}
  \label{fig:results}
  \end{minipage}
\end{figure}

All data preparation, training, and decoding steps completed
successfully. The accuracy rates for evaluation on the testing set are
shown in Figure~\ref{fig:results}.\medskip

\subsection{Performance Assessment}
The model's accuracy in predicting Arabic utterances to be Arabic and
Spanish utterances to be Spanish was $>$99\%, which is higher than the
current state of the art. This high accuracy could be due to a number
of factors, including the inherent robustness of the ASR system used
in predicting the phones or because of high-quality audio and
transcript data among the training and testing sets for these
languages. For Arabic in particular, an additional reason for high
accuracy could be that it is quite distant linguistically and
acoustically from the other three languages, and therefore might be
easier to identify from among these four.\medskip

The model's accuracy in identifying French and Turkish were
significantly lower, though still greater than random choice for all
but African-accented French. The accuracy on the Turkish testing
dataset was about 70\%; one source of error for this dataset in
particular was that the source of the data (not OpenSLR) was quite
obscure since there are very few speech corpora for Turkish, and the
quality of the audio might not be optimal for testing. Alternatively,
there might be an upstream issue with the pronunciation lexicon or
language modeling for Turkish, leading to reduced downstream model
performance. The accuracy of the model in identifying French was about
44\%; this could also be due to the data source, which was part of the
Expressive Speech dataset intended for emotion detection
applications. It is possible that the emotional expression injected
into the the audio could have interfered with identifying the
language. Upon examining the .ctm file containing the phone
predictions for the French utterances, it became apparent that almost
all the phones in every utterance were predicted to be either French
or Spanish, with Spanish in the slight majority. This is unlikely to
be a coincidence; Spanish and French are closely related Romance
languages, and it makes sense that the model might confuse the two
with a bias toward predicting phones as Spanish. Interestingly, this
examination reveals that the ASR system can pick up on linguistic
similarities and relationships between different languages. One
potential followup on this topic could be to build a similar ASR
system that, given a large set of languages, can reconstruct the known
language family tree (e.g tracing the development of various languages
from the larger Indo-European family).\medskip

The prediction accuracy of the model on African-accented French was
about 25\%, which indicates an essentially random prediction. This
result is surprising; the phones in European French and
African-accented French are not that substantially
different~\cite{youtube}, so a predictive model for one French should
apply to the other. It could be that this predictive model is simply
highly sensitive to accent because of over-training or another
technical reason, or perhaps the phone differences are indeed
substantial enough to confound the model.\medskip

\subsection{Further Sources of Error}
An additional factor that could be degrading the performance of this
system is sub-optimal data preparation or network architecture. As
part of the training stage of this system, the training data was
decoded and scored for WER (word error rate; i.e, the error rate of
the model in predicting the correct words in an utterance). At its
lowest, the WER was 54.47\%, which is significantly higher the the
WER reported in the original Tedlium workflow. This could indicate a
number of issues, including not enough training data being provided,
improper processing of the data, inconsistent pronunciation lexicons,
or an incorrect architecture for the neural network.\medskip

\begin{figure}
  \begin{minipage}[b]{1.0\linewidth}
  \centering
  \includegraphics[scale=0.5]{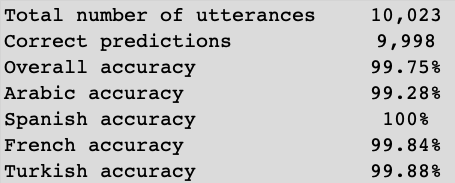}
  \caption{Results of decoding and scoring on the training set.}
  \label{fig:picture1}
  \end{minipage}
\end{figure}

All of the enumerated potentially confounding factors in successful
language identification could be the subject of future followups to
this study in order to improve the system's performance. As an
immediate followup, the training should be redone with only a portion
of the MediaSpeech dataset, with some of the data being withheld for
testing. The language identification accuracy of the model on the
training data is $>$99\% across all languages
(Figure~\ref{fig:picture1}), which could mean that testing on
different data from the same dataset would be more successful, if the
model is not simply over-fitted.\medskip

One concern with the general methodology followed here to train a
language ID model was that the model might learn to identify languages
based on particular characteristics of the training audio rather than
inherent differences in the languages. For example, if all of the
speech data for French were collected with the same microphone and in
the same strictly controlled environment, the model might recognize
the audio as French based on subtle frequencies from that particular
microphone, but fail when presented with French recorded using a
different method of sound capture. Some potential solutions included
artificially adding random noise to the training set for each language
to mitigate distinctive background noise. However, this concern was
obviated in other ways. One, a genuine multilingual dataset
(MediaSpeech) was used for training, so there was increased assurance
of consistent audio quality and recording apparatus across the audio
recordings of each language, reducing the possibility that a
particular language might have wildly distinctive background audio
features. Two, the performance of the model was tested on other
datasets which would not share any distinctive audio features with the
training data. The success of the model in accurately predicting
Arabic and Spanish utterances on external data suggests that the model
has learned real features of the different languages, rather than
simply audio artifacts.\medskip

\subsection{Future Directions}
One way to improve the model and overall performance of the system
could be to drastically reduce the size of the neural network. Since
the language identification task is far more trivial than a precise
speech-to-text transcription task, the 16-layer TDNN used in the
Tedlium recipe is likely unnecessarily large for this purpose. An
overly large network could lead to poor training or over-fitting, and
most certainly leads to increased training times and required
resources. By reducing the number of layers in the network and
otherwise simplifying the architecture, the model would be more
lightweight and train more quickly.\medskip

A more lightweight model could open up additional applications for
this language ID system, including in real-time code-switching
detection. In an environment where one or multiple people take turns
speaking in different languages, a real-time code-switching detection
system could pinpoint the exact times when these switches occur, and
predict the new language to which the conversation has shifted. Since
real-time operation of such a system would require high decoding and
scoring speed, a lightweight model would be ideal.\medskip

\begin{figure}
\begin{minipage}[b]{1.0\linewidth}
\centering
\includegraphics[scale=0.5]{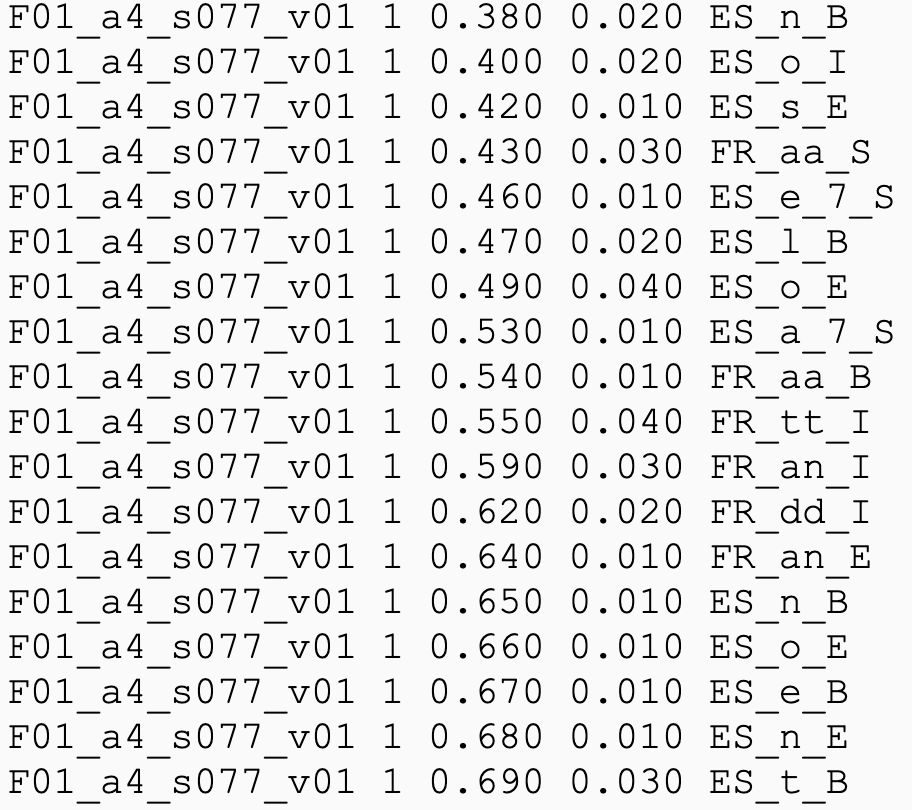}
\caption{Example input to a code-switching evaluation.}
\label{fig:example}
\end{minipage}
\end{figure}

The methodology used in this study could readily transfer over to
building a code-switching detection system. The first step in training
such a system would be to generate artificial data containing multiple
languages in a single utterance, along with timestamps indicating when
the code-switches are known to occur. A similar ASR system to the one
used in this study could be employed; indeed, only the evaluation
stage would need to change significantly. Example input to such an
evaluation script is shown in Figure~\ref{fig:example}. The evaluation
script would read through the first three phones and be confident that
it is detecting Spanish. Although it sees a single French phone next,
it would not decide that the language has switched unless the number
of phones of a different language that it reads is above some
threshold; therefore, this part of the utterance would still be
classified as Spanish. After seeing a few more Spanish phones, the
system would detect a block of French phones and determine that the
language had switched. The time stamps from the third column of this
ctm file would be returned, along with the prediction for the new
language - in this case, French.\medskip

In addition to making the neural network smaller to improve language
identification and code-switching detection performance, the
architecture of the network could also be dramatically altered by
replacing the TDNN model with a simple feedforward network and
softmax. This version would implement language identification not by
reading prepended symbols, but by direct classification. The
disadvantage of this approach is that it would not be as useful for
code-switching detection applications as the original approach, since
it would process whole utterances at a time and would therefore lack
granularity to determine if a language switch has happened in the
middle of the utterance. Nevertheless, alternate schemes for language
ID should be investigated to see if they are indeed more
computationally efficient and accurate.\medskip

\section{Conclusion}
At the outset of this study, a successful language identification
system was defined as one that would achieve an error rate of less
than 10\%. Only the most recent state-of-the-art deep learning methods
have managed to achieve this level of accuracy. Although the approach
used here did not achieve that benchmark for every language, it did
have greater-than-random accuracy in detecting all four
languages. This is encouraging, because it suggests that the approach
set out here is valid and simply needs to be fine-tuned. In subsequent
work the author will indeed fine-tune this method to improve language
identification precision to the greatest extent possible. Using a
greater volume of speech data in training, perhaps sourced from the
multilingual LibriVox or TedX corpora, would likely improve
performance~\cite{Pratap2020MLSAL,salesky2021mtedx}. Interestingly, it
was anticipated that languages with close linguistic relationships
might lead to difficulties in distinguishing between several candidate
languages. The evidence shown in this paper regarding the mistaken
prediction of French phonemes to be Spanish is precisely the expected
scenario. More generally, the accuracy of the approach here will
always likely be higher for some languages than others due to a
variety of factors, including phonetic similarity between languages
and variation in the quality of the recordings. Indeed, the mix of
languages provided in the closed-set for language identification will
also likely make a difference: a closet-set of entirely Romance
languages would likely have poor performance compared to a closed-set
of distant languages. Various aspects of the approach laid out here
will continue to be adjusted in order to eliminate biases and amplify
linguistic differences so as to achieve the highest accuracy
possible.\medskip

\bibliographystyle{IEEEtran}
\bibliography{ms.bib}

\end{document}